\documentclass[10pt,twocolumn,letterpaper]{article}

\usepackage{cvpr}              

\usepackage{graphicx}
\usepackage{amsmath}
\usepackage{amssymb}
\usepackage{booktabs}
\usepackage{tabularx}
\usepackage{multirow}
\usepackage{xcolor}
\usepackage{url}
\author{}
\usepackage{authblk}

\usepackage[pagebackref,breaklinks,colorlinks]{hyperref}

\usepackage[capitalize]{cleveref}
\crefname{sec}{Sec.}{Secs.}
\Crefname{sec}{Section}{Sections}
\Crefname{tab}{Table}{Tables}
\crefname{tab}{Tab.}{Tabs.}

\def\cvprPaperID{61} 
\def\confName{CVPR}
\def\confYear{2022}

\makeatletter
\renewcommand\AB@affilsepx{, \protect\Affilfont}
\makeatother
\title{\vspace{-0.48cm}Bridging the Gap Between Object Detection and User Intent via Query-Modulation\vspace{-0.48cm}}
\date{}
\author[1]{Marco Fornoni}
\author[2,1]{Chaochao Yan}
\author[1]{Liangchen Luo}
\author[1]{Kimberly Wilber}
\author[1]{Alex Stark}
\author[1]{Yin Cui}
\author[1]{Boqing Gong}
\author[1]{Andrew Howard}
\affil[1]{Google Research}
\affil[2]{University of Texas at Arlington}

\begin{document}

\maketitle
\vspace*{-0.48cm}

\begin{abstract}
When interacting with objects through cameras, or pictures, users often have a specific intent. For example, they may want to perform a visual search. 
With most object detection models relying on image pixels as their sole input, undesired results are not uncommon. Most typically: lack of a high-confidence detection on the object of interest, or detection with a wrong class label.
The issue is especially severe when operating capacity-constrained mobile object detectors on-device.
In this paper we investigate techniques to modulate mobile detectors to explicitly account for the user intent, expressed as an embedding of a simple query.
Compared to standard detectors, query-modulated detectors show superior performance at detecting objects for a given user query. Thanks to large-scale training data synthesized from standard object detection annotations, query-modulated detectors also outperform a specialized referring expression recognition system. Query-modulated detectors can also be trained to simultaneously solve for both localizing a user query and standard detection, even outperforming standard mobile detectors at the canonical COCO task.
\end{abstract}

\vspace{-2em}
\section{Introduction}

\begin{figure}[t]
\centering
\includegraphics[width=\columnwidth]{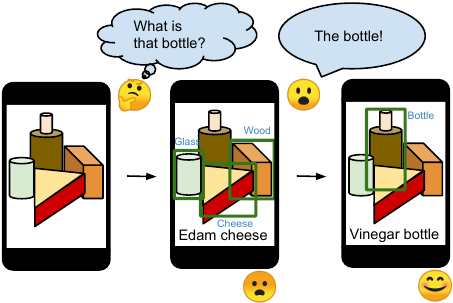}
\caption{Example use case for query-modulated detection: a user points their phone at a group of objects, seeking information about one of them. The model may not detect the object of interest, or it may detect it with the wrong label. 
By explicitly taking into account the user input, and actively searching for the requested object, the model can correct the results, and detect the intended object.
Further along the pipeline, the phone might follow up with nutrition information, or product reviews.}
\label{fig:four_step}
\end{figure}

Convolutional neural networks (CNNs) have transformed computer vision, enabling new use cases, as research pushes the limits of accuracy and efficiency. Still, on capacity-constrained mobile devices even the best performing models can produce results mismatched to the user intent. An example scenario is illustrated in Fig.~\ref{fig:four_step}. 

The goal of this paper is to present a formulation for building mobile object detectors that can leverage user queries to significantly improve detection accuracy.
This new flavor of models and training procedures, named \textit{Query-Modulated object Detectors (QMD)}, 
can even outperform traditional detectors when no query is available, thanks to multitask learning. Computationally, Query-Modulated Detectors are thus more efficient than traditional detection models.

\vspace{1em}
The main contributions of the paper are:\\
1) We introduce the notion of query-modulated detection, the task of adapting an object detector computation to take into account a user query.\\
2) We propose a way to train query-modulated detectors directly using synthetic queries generated from large-scale object detection datasets.\\
3) We present a multi-task training strategy to simultaneously solve for both the  query-modulated detection task, and the traditional object detection task.\\
4) We show that for query-modulated detection, query embeddings can be learned end-to-end, in a lightweight manner, significantly reducing computational and memory costs with respect to traditional visual grounding approaches, depending on large language models such as BERT~\cite{devlin2018bert}.\\
We validate our approach by comparing to standard object detection and referring expression recognition models.



\section{Related Works}
\label{sec:related_works}

In this section we review prior works that most closely relate to our approach, arranging them into two classes: object detection, and referring expression recognition.



\paragraph{Object detection.}
Object detection is one of the fundamental tasks in computer vision, solving for both localizing the objects, and classifying them using a closed set of object labels. Modern object detection models are built as convolutional neural networks and can be broadly divided into two classes---one-stage and two-stage---according to the model architecture.
Two-stage  detection methods \cite{girshick2014rich, ren2015faster, dai2016r, ren2017faster, lin2017feature, he2017mask, cai2018cascade} decompose object detection into object localization and classification by first generating object region proposals, and then refining and classifying those proposals. 
One-stage methods \cite{redmon2016you, Liu_2016, Redmon_2017_CVPR, lin2017focal, law2018cornernet, zhang2018single, redmon2018yolov3, zhao2019m2det, tan2020efficientdet} accomplish localization and classification simultaneously, achieving much lower inference times with some accuracy degradation.
While some of the errors produced by capacity-constrained mobile detectors could be addressed by explicitly taking into account the user intent, architectures capable of doing so~\cite{Yang_2019} have only been explored in the \textit{referring expression recognition} and visual grounding literature.

\paragraph{Referring expression recognition.}
The goal of referring expression recognition is to unambiguously localize an object, or a region in an image referred to by \textit{a natural language expression}, which can be a word, a phrase, a sentence or even a dialogue~\cite{kazemzadeh2014referitgame, plummer2015flickr30k, mao2016generation, yu2016modeling, cirik2018visual, Yang_2019, pont2020connecting}.
Similarly to object detection, existing referring-expression recognition systems employ either two-stage, or one-stage architectures. These are inspired by, or directly derived from, the object detection literature. Two-stage frameworks~\cite{plummer2015flickr30k, mao2016generation, yu2016modeling, wang2018learning, zhang2018grounding, yu2018mattnet} operate by first generating a set of region candidates, and subsequently ranking these regions according to the provided expression. The performance of two-stage frameworks is largely capped by the region proposal generation~\cite{Yang_2019, shrestha2020magnet}, which is not trained simultaneously with the candidates ranking module in an end-to-end manner.
One-stage methods~\cite{yeh2017interpretable, chen2018real, zhao2018weakly, sadhu2019zero, Yang_2019, yang2020improving} recently emerged as the preferred approach. These methods directly regress a bounding box in accordance with the user query. Such solutions typically rely on one-stage detection architectures, such as YOLO~\cite{redmon2016you, Redmon_2017_CVPR, redmon2018yolov3} or SSD~\cite{Liu_2016}.

Referring Expression recognition is currently considered expensive. First, data acquisition is costly: for each given object of interest, multiple referring expressions need to be collected and validated by different annotators~\cite{kazemzadeh2014referitgame}. Second, computationally expensive NLP models such as BERT~\cite{devlin2018bert} need to be used to process the verbal expressions. On the other hand, \cite{kazemzadeh2014referitgame} reports that \textit{50\% of the referring expressions in their study is composed only of a noun}, while \textit{82\% of the remaining ones uses only one additional attribute}, most typically \textit{the object coarse location} (e.g.\ left, right, bottom). In other words, \textit{91\% of the objects can be correctly localized with referring expressions as simple as an object label, plus an optional coarse location}.
A basic referring expression recognition system could thus be directly trained on existing large-scale object detection datasets, by synthesizing the referring expressions from the groundtruth bounding-box annotations, and stripping all NLP processing. Such an approach could be built on top of high-performance object-detection training pipelines, and scaled through inexpensive data annotation pipelines.  In other words, \textit{it'd come almost for free}.





\begin{figure*}[ht]
\centering
\includegraphics[width=\textwidth]{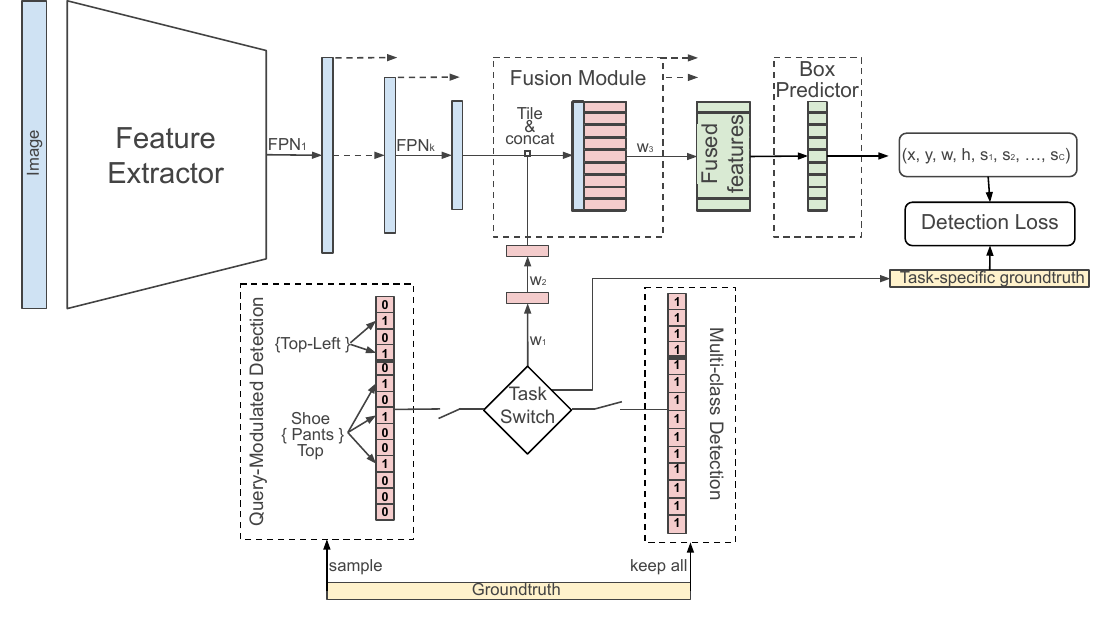}
\caption{The Query-Modulated Detector (QMD) architecture, and its multi-task training pipeline. The reserved $[1, 1,\ldots , 1]$ query triggers standard object detection. All other values are treated as detection queries. A query is represented by a $k$-hot encoding, and is divided in two parts: the label encoding, and the location encoding. During training the model alternates between detecting the objects referred by randomly synthesized queries, and standard object detection.}
\label{fig:multi_task_approach}
\end{figure*}
\section{Approach}\label{sec:approach}
In this paper we formulate the problem of referring-expression recognition in the context of object detection, and refer to it as \textit{Query-Modulated Detection}. We focus on simple yet effective queries, consisting of an object label, plus an optional coarse location (e.g.\ ``top-right''). 
Similarly to referring expression recognition, we assume that when a user provides a given query, the target object must be present in the image. 
In contrast to referring expression recognition models:  1) We directly train on \textit{synthetic queries} from common object detection datasets. 2) We employ a \textit{multi-task training} strategy to optimize simultaneously for both standard object detection and query-modulated detection. 3) Since we focus on a closed vocabulary with a limited set of words (the set of labels supported by the detector), we optionally replace BERT embeddings with \textit{inexpensive K-Hot binary encodings}, and learn the query embedding end-to-end.
Furthermore,  while visual referring expression recognition is traditionally a single-object localization problem (there exists one and only one image region corresponding to the expression), our setting is slightly more challenging, as we ask the model to detect \emph{all} the objects corresponding to the given query


The architecture for our model is shown in Fig.~\ref{fig:multi_task_approach}. It features two inputs: the image (in blue), and the query encoding (in red). Similarly to~\cite{Yang_2019}, an embedding for the query is computed by two fully-connected layers ($w_1$ and $w_2$), followed by a $\ell_2$-normalization. The query embedding is then tiled and concatenated to the last convolutional feature maps (also $\ell_2$-normalized), before the box predictor. A $1\times1$ convolution is then employed to fuse the query with the image features. Finally, the fused feature maps are passed to a canonical SSD box predictor~\cite{Liu_2016}.


\subsection{Query Synthesis}
\label{sec:synthetic_query_generation}
As discussed in Sec.~\ref{sec:related_works}, 91\% of the objects in the ReferIt dataset~\cite{kazemzadeh2014referitgame} can be correctly localized with referring expressions as simple as the object label, plus an optional coarse location.
We thus propose to generate simple synthetic referring expressions from standard object detection groundtruth annotations, provided by large-scale datasets such as Open Images~\cite{OpenImages} and COCO~\cite{lin2014microsoft}.
Specifically, we focus on referring expressions containing the categorical label of the object, plus an optional coarse spatial location (e.g.\ ``on the right'').
Differently from the referring expression recognition literature, we ask our models to detect \textit{all} objects matching a given expression, as in our case more than one object may be associated to the expression. We also allow for queries containing multiple object labels at the same time, e.g.\ all clothing labels, or all vehicles labels. For brevity, we use the term \textit{query} to indicate such loose referring expressions, and use the term \textit{``query-modulated detection''} to indicate the task.
We focus on the following three types of queries, which can be directly synthesized and evaluated on object-detection datasets:
\begin{itemize}
    \item \textit{Single-Label Detection (SLD)}. Detect all objects for a single label, sampled from the image groundtruth.
    \item \textit{K-Label Detection (KLD)}. Detect all objects for $K$ random labels, obtained by independently sampling (with probability 0.5) each label from the image groundtruth.
    \item \textit{Localized-Label Detection (LLD)}. Detect all objects for a single random label, and a coarse spatial location. Both the label and the coarse location are sampled from the image groundtruth.
\end{itemize}

\subsubsection{Localized-Label Detection Query Synthesis}
In this section we provide the details of the approach used to synthesize LLD queries. Based on the spatial distribution in~\cite{kazemzadeh2014referitgame}~(Fig.~3), and to reflect the way users typically refer to spatial locations, we employ a few predefined slices for each axis of a standard $4\times4$ grid. As shown in Fig.~\ref{fig:spatial_grid}, we use 3 reference overlapping slices for the $y$-axis: \{\textit{top}, \textit{center}, \textit{bottom}\}, and 5 partially overlapping slices for the $x$-axis: {\textit{far-left}, \textit{left}, \textit{center}, \textit{right}, \textit{far-right}\}. For brevity, we refer to those as $y$-slices and $x$-slices. Furthermore, for each axis we also employ an additional \textit{all} slice, to indicate a lack of constraint on that axis. A location constraint is finally defined as the product of a $y$-slice constraint, and a $x$-slice constraint. E.g.\ (top,~right), (bottom,~left), (all,~left). 

As in~\cite{kazemzadeh2014referitgame}, we assume the user would provide a location constraint only if it is necessary to correctly identify the target object. For example, if the user wants to indicate the car in the image, and the image contains only one car instance, location constraints are not necessary to correctly detect the desired object.
Furthermore, we assume that the user will select the spatial constraint that most clearly identify the object of interest. For example, if both \textit{far-right}, \textit{right}, and \textit{all} are valid $x$-slices, we assume the user would pick the most informative one, namely: \textit{far-right}.

More formally, we adopt the following approach to synthesize LLD queries from the image groundtruth:
\begin{enumerate}
    \item Randomly select a target label, among the labels in the image groundtruth, and prune out all boxes not from this label.
    \item If the target label contains more than one object instance in the image:
    \begin{enumerate}
        \item Randomly select one of the instances as the final target.
        \item Select the tightest ($y$-slice,~$x$-slice) constraint containing the target instance. In case of multiple possible solutions, select the one containing the least number of groundtruth boxes. E.g.\ if the target object lies in the overlap between the \textit{center}, and \textit{right} $x$-slices, select the one containing the least number of groundtruth boxes.
    \end{enumerate}
    \item Finally, prune all remaining boxes not contained in the selected ($y$-slice,~$x$-slice) constraint .
\end{enumerate}

\begin{figure}[ht]
\centering
\includegraphics[width=.75\linewidth]{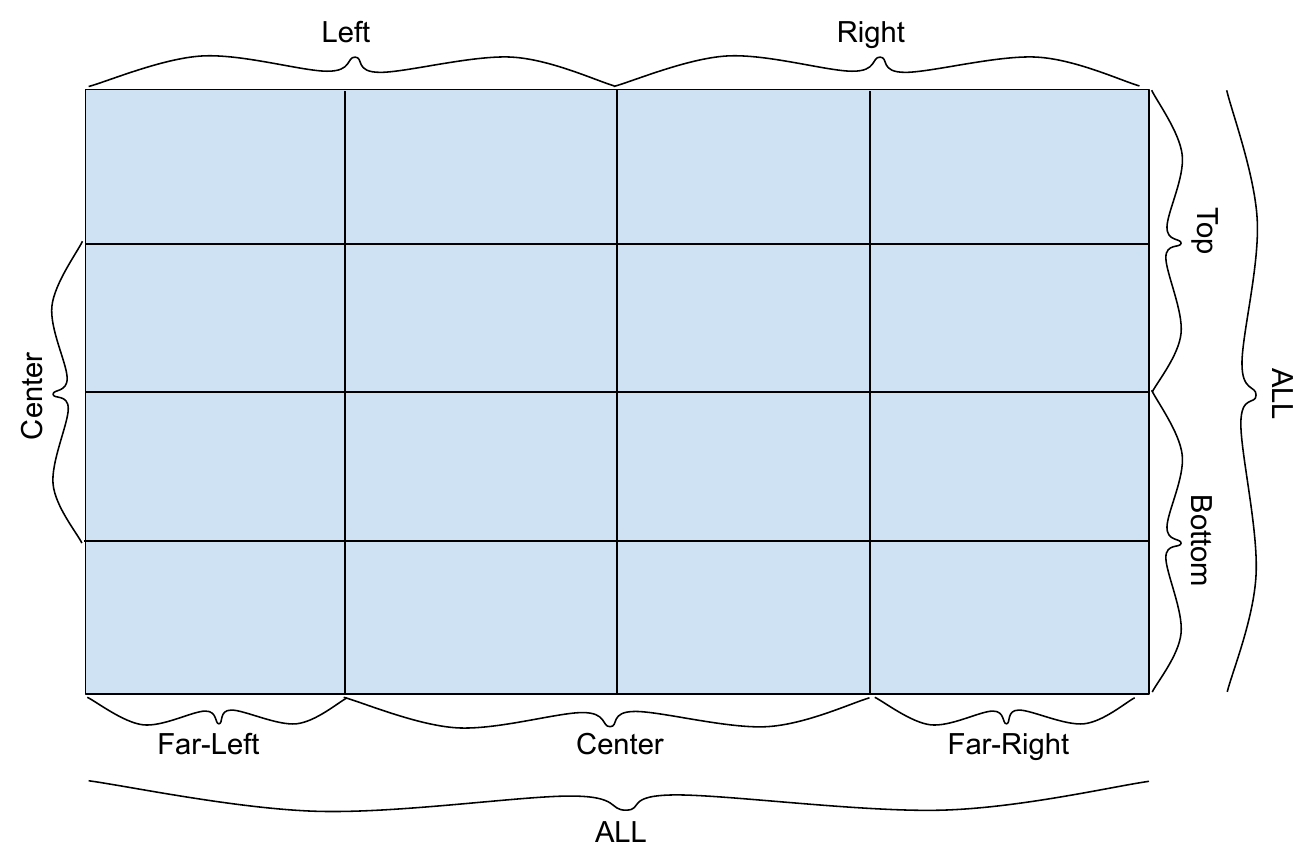}
\caption{Grid and reference $x$-slices (horizontal axis), and $y$-slices (vertical axis). A $4\times4$ grid is divided into 3 overlapping slices on the $y$ axis, and 5 overlapping slices on the $x$ axis. Additional two slices fully covering each axis are used to denote lack of constraints on that axis.}
\label{fig:spatial_grid}
\end{figure}

We use $\mathit{IntersectionOverArea} \geq 0.9$ as the criteria for considering a box as contained in a cell grid. Namely the intersection between the box and the grid, divided by the box area should be at least 0.9.


\subsection{Query encoding}
\label{sec:query_encoding}
We represent user queries using \textit{$k$-hot encodings}. For label encoding, given an object detection problem with $C$ classes, we use a $C$-dimensional binary vector, where each bit indicates a given class label, and a value of 1 indicates that all objects from that class need to be localized. For location encoding, we again use a binary representation. We use the first 3 bits to represent the $y$-slice, and the next 5 bits to represent the $x$-slice, and the all-ones encoding to represent the lack of constraints on the $y$-axis, or the $x$-axis.  For example, (top,~right) is represented by the [1,0,0,0,0,0,1,0] encoding, (all,~far-right) is represented by the encoding [1,1,1,0,0,0,0,1], and the complete lack of constraints (all,~all) is represented by the encoding: [1,1,1,1,1,1,1,1].
The final k-hot encoding is obtained by concatenating the location encoding, to the label encoding.

With respect to BERT embeddings~\cite{devlin2018bert}, $k$-hot encodings are extremely efficient to compute and store, and  do not introduce any bias in how they model classes. BERT embeddings, on the other hand, incorporate extra prior knowledge, since similar class names are supposed to have a closer distance in embedding space.
We experimentally compare $k$-hot encodings with BERT~\cite{devlin2018bert} embeddings in Sec.~\ref{sec:ablation_1hot_vs_khot}.


\subsection{Multi-Task Training}
\label{sec:multi_task_multi_class_qml_detector}

A model trained solely to perform query-modulated detection is not suitable as a stand-alone replacement of a standard object detector (see Sec.~\ref{sec:single_task_query_modulated_detector}). 
The main reason is that the model is trained only using queries for objects actually present in the image. For example, when presented with an image not containing the queried label, the modulated feature maps may ``hallucinate'' the desired object, returning high-confidence boxes on unrelated objects. To address the above issue, we introduce a new multi-task training strategy. Specifically, for each training image we randomly switch, with a given task switching ratio, between:
\begin{enumerate}
    \item Query-modulated detection. Using a query synthesized and encoded as described in~\ref{sec:synthetic_query_generation}.
    \item Standard object detection. In this case the model is provided the fixed $[1, 1, 1,\ldots , 1]$ query, activating all labels regardless of the image groundtruth.
\end{enumerate}
We call this approach the \textit{Query-Modulated object Detector (QMD)}. The full architecture is depicted in Fig.~\ref{fig:multi_task_approach}. Experimental results are provided in 
Sec.~\ref{sec:experiments}.
\section{Experiments}
\label{sec:experiments}
\subsection{Implementation details}
We implemented our models in TensorFlow Object Detection API~\cite{huang2017speed}, using a RetinaNet~\cite{lin2017focal} architecture with both mobile and server-side backbones. For the mobile-friendly models we employed a MobileNet-v2~\cite{Sandler_2018}, with 1.0 width multiplier, 320px resolution, L3-L7 128-d FPNLite, and a fully convolutional box predictor with: weight-sharing across scales, four 128-d layers, and depthwise-separable convolutions. For the server models we employed a ResNet-101~\cite{He_2016}, with 1024px resolution, L3-L7 256-d FPN, and a fully convolutional box predictor with four 256-d layers per scale.
For QMD, the query encoding is first passed through 2 fully connected layers with 512 neurons each, then $\ell_2$-normalized, tiled and concatenated across the spatial dimension of the $\ell_2$-normalized FPN feature maps. Finally, the concatenated features are fused together using a $1\times1$ convolution, and fed to the box predictor.

\subsection{Datasets}
The ReferIt dataset~\cite{kazemzadeh2014referitgame} comprises 19,894 images, annotated with 130,525 expressions spanning 96,654 distinct objects. We employ the standard 9,000, 1,000 and 10,000 split from~\cite{Hu_2016} for training, validation, and test respectively. The COCO dataset~\cite{lin2014microsoft} is a 80-class common object detection dataset. We use the 2017 version of the dataset. Since the test annotations are not released and our evaluation protocol differs from the official one, we perform all our evaluations on the validation set.
Open Images Detection (OID) v4~\cite{OpenImages} is a large-scale, large-vocabulary dataset, composed of more than 1.7M training, 40K validation, and 125K test images, spanning 600 different classes. Since the test set annotations are made available, for this dataset we report metrics as measured on the test set (except where specified).

\subsection{Metrics}
To evaluate accuracy we adopt the following metrics:
\begin{enumerate}
    \item Detection (DET): standard mAP for object detection.
    \item Query-modulated detection. In this setting the model is asked to detect all objects for a given query. The image is guaranteed to contain at least one object matching the query. We report the AP measured when solving for SLD, KLD, or LLD.
\end{enumerate}
Since for query-modulated detection the target label is provided as an input to the model, it is superfluous to evaluate the model labeling accuracy. After pruning the groundtruth with the desired label, we thus assign the remaining groundtruth and all detected boxes to one single class-agnostic label, and evaluate the predictions in a class-agnostic fashion. Following the standard protocol, COCO (m)AP is computed using the standard set of [.5:.95] IOUs, while Open-Images (m)AP is computed with the Open Image V2 metric~\cite{oid_V2_detection_metrics} using 0.5 IOU.

\subsection{ReferIt and post-processing baselines for SLD}
\label{sec:referit_and_post_processing_baselines}
\textbf{Why not just using a referring expression model?} As a first experiment to motivate our approach, we use the COCO 2017 validation set as  the target dataset, and compare two obvious baselines to perform query-modulated detection. Namely: 1) Evaluating a model specifically trained for general referring expression recognition~\cite{Yang_2019}; 2) Post-processing a standard object detector to retain only the boxes specified by the user query.

For this experiment, we focus on the single label detection (SLD) metric, as object labels are the most important and common form of referring expressions~\cite{kazemzadeh2014referitgame}, and correctly recognizing them is a necessary condition for recognizing more complex ones.

As a referring expression recognizer baseline, we re-implement the One-Stage BERT referring expression recognition model~\cite{Yang_2019}, and train it on the ReferIt~\cite{kazemzadeh2014referitgame} dataset. Specifically, we use a one-stage SSD architecture with a ResNet-101 backbone at 320px, with query fusion and spatial encoding features, as in~\cite{Yang_2019}. For query embedding we use a 768-d BERT embedding from the CLS token. Also, similarly to~\cite{Yang_2019} we employ several augmentations to avoid over-fitting on this small training set: random brightness, hue, saturation, contrast, gray-scale and random crop augmentations. We use a COCO pre-trained feature extractor and froze all layers, only training the fusion and box predictor modules. Since we are aiming at evaluating the model on detection data, we neither add a softmax non-linearity across all anchors, nor do we optimize the anchors on the ReferIt dataset (as done by ~\cite{Yang_2019}).  This model achieves 57.1\% top-1 accuracy on ReferIt~\cite{kazemzadeh2014referitgame}. This is slightly inferior\footnote{The delta is possibly explained by the absence of soft-max across anchors, the lack of anchors customization, as well as the lack of fine-tuning of the feature extractor.} to the 59.3\% reported in~\cite{Yang_2019}, but far above the other approaches benchmarked in~\cite{Yang_2019}, so we feel it is a comparable reproduction.

For the second baseline, we train a standard ResNet101-SSD detector on COCO (Row~5 in \cref{tab:single_task_qmd_results}) and post-process its outputs by pruning all detections except for that of the query class.
\begin{table}
\footnotesize
    \centering
    \begin{tabularx}{\linewidth}{|X p{1.5cm} p{1.5cm}|}
    \hline
    {Baseline} & {SLD AP} & {SLD AR@1} \\
    \hline
    {Object Detector + Post-processing} & \textbf{47.72} & \textbf{26.32} \\
    {ReferIt Model~\cite{Yang_2019}} & 12.19 & 12.23 \\
    \hline
    \end{tabularx}
\caption{\label{tab:referit-baseline}We compare the performance of a One-Stage BERT Referring Expressions model~\cite{Yang_2019} trained on ReferIt~\cite{kazemzadeh2014referitgame} to a simple post-processed object detector, on the subset of the COCO validation that only contains ReferIt entities (labels). See text for details.}
\end{table}

For a fair evaluation, we only use the subset of the COCO validation set containing the 73 COCO labels that also appear in the ReferIt vocabulary (everything except \textit{baseball bat}/\textit{glove}, \textit{fire hydrant}, \textit{hair drier}, \textit{hot dog}, \textit{parking meter}, \textit{tennis racket}). This corresponds to 4,926 validation images.

Results are shown in \cref{tab:referit-baseline}.
The post-processed object detector handily outperforms the ReferIt model in both SLD AP and Recall@1. Besides the ReferIt training set being much smaller than COCO (9,000 images vs 118,287) and the dataset shift, this could possibly due also to the ReferIt model being trained with only a single box associated to each query. This last factor can be excluded by noting that Average Recall @ 1 is also much lower with respect to the COCO model. This metric considers only one single top-confident box for each (query, image) pair.\footnote{For example, if the dataset contained only one image with three ground-truth ``car'' boxes, and for the ``car'' query the ReferIt model predicted only one box on one single car, while the COCO model predicted three boxes on three cars, the AR@1 for the two models would be identical.}

%
%

\emph{To summarize, ReferIt models have poor generalization abilities due to the very limited number of training samples, and severely under-perform a simple post-processing baseline. This motivates developing query-modulated detectors that can directly be trained on large-scale detection datasets.}

\subsection{Single Task Query-Modulated Detector}
\label{sec:single_task_query_modulated_detector}
In this Section we discuss the results achieved by QMD trained solely to solve for query-modulated detection. We consider both mobile and server models, and perform experiments on both COCO-17 and Open~Images~Detection~v4, for both object-detection, and query-modulated detection.
\begin{table}[h]
\footnotesize
    \centering
    \begin{tabularx}{\linewidth}{|p{0.5cm}p{1.2cm}p{0.8cm}|p{1.0cm}|p{1.0cm}|X|}
        \hline
        Dataset & Backbone & Model & SLD AP & KLD AP & DET mAP\\
        \hline
        OID & MobileNetV2 & Detector & 51.3 & 45.4 & \textbf{27.3} \\
        OID & MobileNetV2 & QMD & \textbf{67.6} & \textbf{57.2} & 1.4 \\
        \hline
        COCO & MobileNetV2 & Detector & 28.9 & 25.6 & \textbf{22.6} \\
        COCO & MobileNetV2 & QMD & \textbf{33.4} & \textbf{28.0} & 9.8 \\
        \hline
        COCO & ResNet101 & Detector & 47.7 & 44.0 & \textbf{38.9} \\
        COCO & ResNet & QMD & \textbf{49.6} & \textbf{46.1} & 22.7 \\
        \hline
    \end{tabularx}
    \caption{\textbf{Single-Task results on OID and COCO}. For SLD / KLD models are asked to put boxes on objects belonging to specific label(s) in each image. Labels are selected among those appearing in the groundtruth. For the standard detector this is obtained by pruning all detections from labels other than the requested one(s).}
    \label{tab:single_task_qmd_results}
\end{table}
In \cref{tab:single_task_qmd_results} we report the SLD AP, KLD AP and detection mAP achieved by these models on the the OID test set, and the COCO validation set. As a baseline, we report the performance of a standard object detector using the same architecture hyper-parameters (feature extractor, input resolution, FPN layers, etc.).
To solve for SLD, and KLD using a standard object detector, predictions are pruned to retain only those matching the classes specified in the query. All detections are then mapped to a single class. 

To evaluate the ability of QMD to operate also as a standard object detector we also compare the COCO mAP achieved on the task of detecting all objects in the image. For QMD this is achieved by activating all labels on all images, using a $[1,1,1,\ldots , 1]$ query. 
By modulating the FPN activations to focus on the queried objects, the MobileNet-based QMD is able to outperform the object detector by +16.3\% SLD AP on OID, and +4.5\% SLD AP on COCO. The ResNet model still improves COCO SLD AP by +1.9\%.
On the other hand, in the standard detection setting, triggered by the $[1,1,1,\ldots , 1]$ query, the single-task QMD severly underperforms the standard object detector.

The main advantage of QMD with respect to post-processing a standard object detector is that QMD is 
trained for actively ``searching'' the desired object(s) in the image. This is particularly important for large-vocabilary datasets like OID, where the box classification problem is considerably harder compared to COCO, and post-processing might not be enough to confidently and stably surface the objects belonging to the user-requested class.

\textit{To summarize, QMD significantly outperforms the post-processing baseline for SLD and KLD. This is especially true for large-vocabulary problems, where class-confusion is a more important issue. However, single-task training is not directly suitable for solving standard object detection.}


\subsection{Multi-Task Query-Modulated Detector}\label{sec:experiments_multi_class_query_modulated_detector}
\begin{figure*}[ht]
\centering
\includegraphics[width=0.9\linewidth]{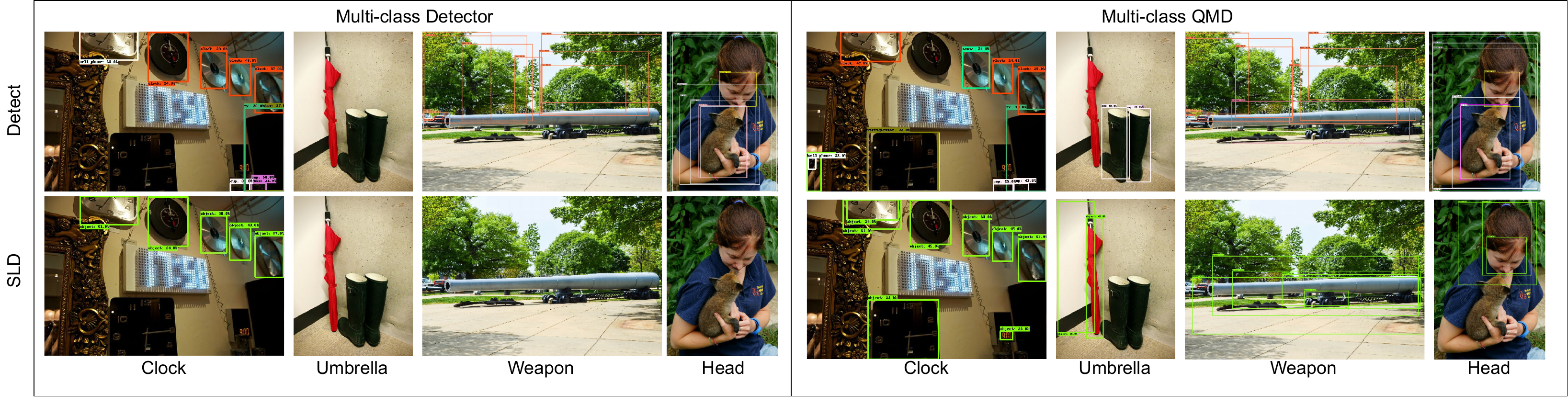}
\caption{Eight top-scoring detections above 0.2 confidence, on COCO validation (clock, umbrella) and OID test (weapon, head) images. Left: results using a standard object detector. Right: results with Multi-Task QMD. First row: object detection results. Second row: SLD results, with the corresponding query.}
\label{fig:multiclass_vs_qmd}
\end{figure*}
In this Section we analyze the performance achieved by QMD trained for both query-modulated detection and standard object detection, as described in \cref{sec:multi_task_multi_class_qml_detector}.
\begin{table}[h]
\footnotesize
    \centering
    \begin{tabularx}{\linewidth}{|p{0.5cm}p{1.2cm}p{0.8cm}|p{1.0cm}|p{1.0cm}|X|}
        \hline
        Dataset & Backbone & Model & SLD AP & KLD AP & DET mAP\\
        \hline
        OID & MobileNetV2 & Detector & 51.3 & 45.4 & 27.3 \\
        OID & MobileNetV2 & QMD & \textbf{63.1} & \textbf{51.5} & \textbf{27.4} \\
        \hline
        COCO & MobileNetV2 & Detector & 28.9 & 25.6 & 22.6 \\
        COCO & MobileNetV2 & QMD & \textbf{32.0} & \textbf{27.8} & \textbf{23.3} \\
        \hline
        COCO & ResNet101 & Detector & 47.7 & 44.0 & \textbf{38.9} \\
        COCO & ResNet & QMD & \textbf{50.0} & \textbf{46.5} & 38.6 \\
        \hline
    \end{tabularx}
    \caption{\textbf{Multi-Task results on OID and COCO}. QMD matches Detector DET mAP, while outperforming on SLD and KLD AP.}
    \label{tab:multi_task_qmd_results}
\end{table}

\vspace{-1em}
Results for OID-v4 and COCO-17 are summarized in \cref{tab:multi_task_qmd_results}. With multi-task training QMD is able to match the standard detector DET mAP, while still providing large gains on SLD / KLD (+11.8\% on OID). As shown in Fig.~\ref{fig:multiclass_vs_qmd}, by conditioning on the desired labels, QMD is able to detect the desired objects even when they are missed, or wrongly labelled when using the model as a multi-class detector.
For a per-class analysis of the results, please refer to the supplementary material.

\textit{To summarize, it is possible to augment an object detector with the ability of leveraging user queries to significantly improve SLD and KLD AP for the desired labels, while preserving the DET mAP when no query is specified.}

\subsection{Efficiency Analysis}
\label{sec:efficiency_analysis}
In this Section we provide an analysis of QMD efficiency compared to post-processing a standard object detector. Similarly to~\cite{tan2020efficientdet}, for both QMD and the standard detector baseline we simultaneously scale up the model width, depth, and resolution. When scaling the model width, we increase the width multiplier for backbone, FPN, and box predictor. When scaling the model depth we increase the number of layers in the box predictor. Specifically, with $D \in \{0, 1, 2, 3, 4, 5, 6, 7, 8\}$, we set:
\begin{itemize}
    \item $width\_multiplier=1.22^D$
    \vspace{-0.5em}
    \item $layers(box\_predictor)=4+D$
    \vspace{-0.5em}
    \item $resolution=320+64*D$
\end{itemize}
$D=0$ corresponds to models in \cref{tab:multi_task_qmd_results}. For each subsequent $D$ value, FLOPs are approximately doubled  (Fig.~\ref{fig:accuracy_vs_flops_coco}).
\begin{figure}[ht]
\centering
\includegraphics[width=0.99\linewidth]{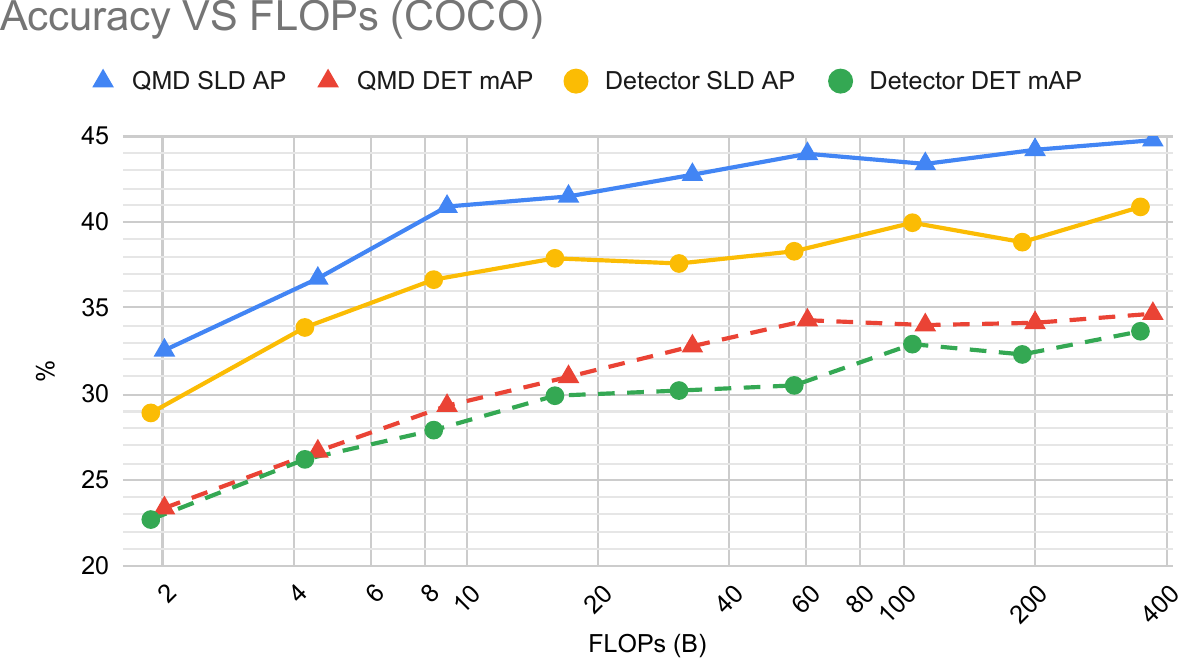}
\caption{COCO SLD AP and DET mAP for QMD and a standard Detector, when varying FLOPs. The FLOPs axis is in log-scale.}
\label{fig:accuracy_vs_flops_coco}
\end{figure}

We observe that:
1) QMD FLOPs overhead WRT standard detectors is between 6.8\% (372.7B vs 349B - slowest model) and 7.4\% (2.02B vs 1.88B - fastest model). 
2) For all considered computational budgets, QMD outperforms the vanilla-detector on both SLD AP \textit{and} DET mAP.  We attribute the DET task efficiency gains to the SLD task positively affecting the DET task during multi-task training.
3) QMD achieves the same SLD AP of the largest vanilla-detector (40.9\%), with 39x less FLOPS (9B vs 349B).

\textit{To summarize, QMD is largely more efficient compared to traditional object detectors for the SLD task. Thanks to multi-task training, it is also equally or more efficient for the standard COCO DET task.}

\subsection{Localized Label Detection}\label{sec:experiments_lld}
In this Section we report the performance of QMD when trained to perform Localized Label Detection (LLD). For this experiment we only use MobileNetV2 backbones and, as for the ReferIt model in \cref{sec:referit_and_post_processing_baselines}, we use the spatial encoding from~\cite{Yang_2019}.
\begin{table}[h]
\footnotesize
    \centering
    \begin{tabularx}{\linewidth}{|p{0.5cm}p{1.0cm}|p{1.0cm}|p{1.0cm}|p{1.0cm}|X|}
        \hline
        Dataset & Model & LLD AP & SLD AP & KLD AP & DET mAP\\
        \hline
        OID & Detector & 55.4 & 51.3 & 45.4 & 27.3 \\
        OID & QMD & \textbf{68.1} & \textbf{62.5} & \textbf{50.0} & \textbf{27.4} \\
        \hline
        COCO & Detector & 31.1 & 28.9 & 25.6 & 22.6 \\
        COCO & QMD & \textbf{35.9} & \textbf{32.4} & \textbf{28.0} & \textbf{23.5} \\
        \hline
    \end{tabularx}
    \caption{\textbf{Localized Label Detection results on OID and COCO}. All models use a MobileNetV2 backbone. QMD is trained in a multi-task fashion, to solve for both LLD and DET.}
    \label{tab:lld_qmd_results}
\end{table}
As reported in \cref{tab:lld_qmd_results}, adding location constraints further improves the results for all models with respect to SLD. This is expected, as we are making the problem even simpler by constraining the area where the objects of interest are searched. Still, similarly to \cref{sec:experiments_multi_class_query_modulated_detector}, the improvement from SLD to LLD is larger for QMD than for Detector (\textbf{+5.6\%} $|$ \textbf{+3.5\%} on OID $|$ COCO for QMD, vs +4.1\% $|$ +2.2\% for Detector), even though QMD SLD baseline is higher (62.5\% $|$ 32.4\% on OID $|$ COCO, vs 51.3\% $|$ 28.9\% for Detector). 
Visualizations for the localized queries and the corresponding results are provided in the supplementary material.

\emph{To summarize, while coarse location constraints can be used to further improve accuracy for all models, QMD is more effective at leveraging them.}

\subsection{Ablation studies}
We provide here several ablation studies, analyzing the impact of model and training hyper-parameters. All experiments in this section are performed on the COCO and OID validation sets.

\paragraph{SLD vs KLD training.}\label{sec:ablation_1hot_vs_khot}
We consider two different QMD training policies, corresponding to the SLD, and KLD metrics. In the SLD case, training queries are built by sampling one single label from the image groundtruth, with embeddings build as $1$-hot vectors. In the KLD case, training queries are built by sampling each groudtruth label with a probability of $0.5$, with embeddings built as $k$-hot vectors. All experiments for this study are performed using only the MobileNetV2 backbone, and only on the COCO 2017 validation set. To minimize noise, we only perform this ablation on the single-task QMD model.
\begin{table}[h]
\footnotesize
    \centering
    \begin{tabularx}{\linewidth}{|p{2.6cm}|X|X|X|}
        \hline
        Training & SLD AP & KLD AP \\
        \hline
        SLD ($1$-hot) & 33.3 & 20.0 \\
        KLD ($k$-hot) & \textbf{33.4} & \textbf{28.7} \\
        \hline
    \end{tabularx}
    \caption{\textbf{Single-Task MobileNetV2 QMD AP, training with $1$-hot and $k$-hot, on COCO 2017}.
    All evaluation results are reported on the validation set.}
    \label{tab:k_hot_vs_one_hot}
\end{table}

Results in \cref{tab:k_hot_vs_one_hot} show that KLD-training largely outperforms 
SLD for the KLD task, while also matching or outperforming SLD-training for the SLD task. We thus adopt KLD as the standard training for all experiments in \cref{sec:experiments_multi_class_query_modulated_detector,sec:efficiency_analysis,sec:experiments_lld}.

\paragraph{BERT embeddings.} We analyze how BERT~\cite{devlin2018bert} embeddings compare to binary $1$-hot / $k$-hot embeddings. As in \cref{sec:referit_and_post_processing_baselines}, BERT embeddings are computed by passing the textual label to BERT, and extracting the CLS token. KLD BERT embeddings are obtained by \textit{averaging} the BERT embeddings of the $k$ labels.

\begin{table}[hb]
\footnotesize
\centering
\begin{tabular}{|r|r|r|r|r|}
\hline & \multicolumn{4}{c|}{Training/Evaluation} \\ 
\cline{2-5} 
BERT Type & SLD/SLD & SLD/KLD & KLD/SLD & KLD/KLD \\ \hline
Mobile (192) & \textbf{33.4} & 13.1 & \textbf{33.0} & 28.0 \\
\cline{1-5} 
Base (768) & \textbf{33.4} & \textbf{18.6} & 32.7 & 27.9 \\
\cline{1-5} 
Large (1024) & 33.3 & 18.4 & 32.8 & \textbf{28.3} \\
\hline
\end{tabular}
\caption{\textbf{Single-Task QMD results using BERT embeddings.} For KLD, BERT embeddings of the $k$ labels are simply averaged.}
\label{tab:bert_vs_hard_coding}
\end{table}

In \cref{tab:bert_vs_hard_coding} we provide detailed results for different mobile and server BERT embeddings. Increasing the embedding size from 192 to 768, or 1024 does not significantly improve results. 
Furthermore, comparing results in \cref{tab:bert_vs_hard_coding} with those in \cref{tab:k_hot_vs_one_hot} shows that in our closed-world object-detection settings, binary $1$-hot / $k$-hot embeddings achieve similar or better performance than BERT embeddings. Please note that \textit{binary $1$-hot / $k$-hot embeddings are much cheaper to compute and store} with respect to their BERT counterpart.
Based on the above observations we employ binary $1$-hot / $k$-hot embeddings throughout the paper.


\paragraph{Detection Task Sampling Ratio.}\label{sec:multi_task_det_ratio}
\begin{figure}[ht]
\centering
\includegraphics[width=\linewidth]{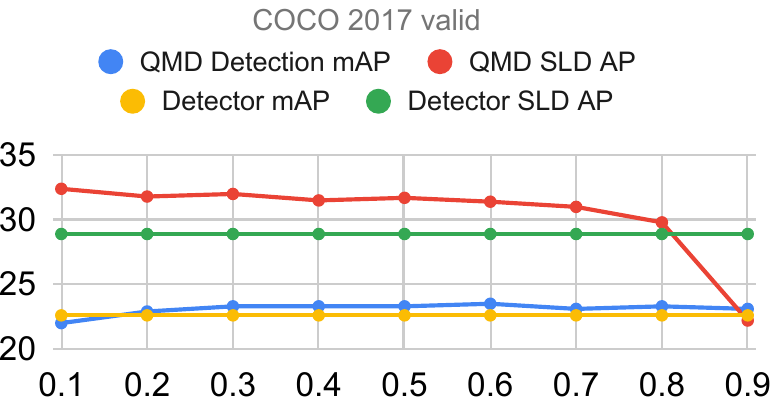}
\vspace{0.5em}
\includegraphics[width=\linewidth]{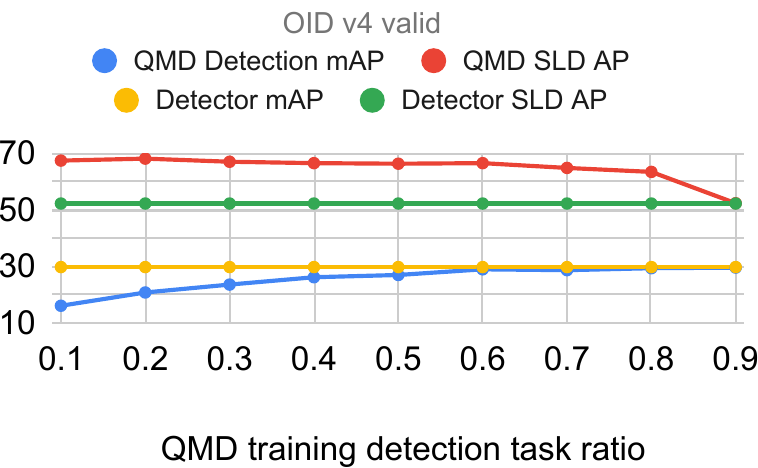}
\caption{Effect of varying the training detection task sampling ratio on the COCO 2017 and OID v4 validation sets.}
\label{fig:detection_task_training_ratio}
\end{figure}
\vspace{-1em}
Figure~\ref{fig:detection_task_training_ratio} shows the effect of varying the detection task sampling ratio during training, for both COCO and OID validation sets. The optimal value is dataset dependent. For both datasets and most choices of the task ratio QMD SLD AP is largely better than Detector SLD AP. On the COCO dataset, multi-task training improves also the DET mAP for most values.


\section{Conclusions}
\label{sec:conclusions}
In this work we presented a formulation to build object detectors that can be queried for detecting specific objects of interest. We focused on simple queries containing only the class label(s) of the object(s) of interest, and optionally a coarse location. We described how to synthesize and encode such queries from standard object detection annotations. We demonstrated that a ReferIt model does not generalize well to large-scale object detection problems, and is outperformed by a simple detector plus post-processing baseline. We showed how the post-processing baseline is in turn largely outperformed by Query-Modulated Detectors. This is particularly true on large-vocabulary datasets, where class-confusion is a more severe issue. We also showed how by jointly training for both standard object-detection and query-modulated detection, one can efficiently and simultaneously solve both problems. Thanks to multi-task training, QMD even improves performance on the original COCO object-detection task.
Finally, we showed how for QMD, a simple $k$-hot query encoding performs equally to BERT, while being much cheaper to compute and store.
Our formulation is generic and can potentially support other types of queries, and query embeddings. In the future we plan to investigate training QMD on more complex queries, still synthesized from large-scale object detection datasets.

{\small
\bibliographystyle{ieee_fullname}
\bibliography{main}
}

\clearpage

\newpage

{\LARGE Supplementary Material}

\setcounter{section}{0}

\section{Visualizations for Localized-Label Detection}
\label{sec:lld_visualizations}
\begin{figure*}[ht]
\centering
\includegraphics[width=\textwidth]{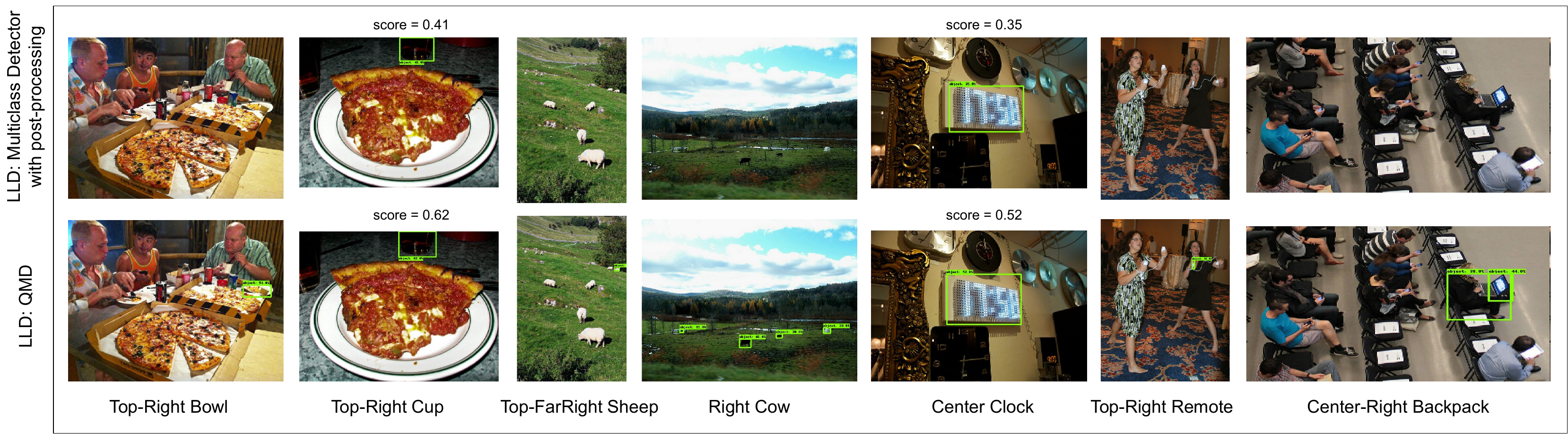}
\caption{Visualization of results for the MobileNetV2 models with Localized Label Detection (LLD) queries. Top: results obtained by post-processing a standard object detector. Bottom: results obtained by QMD. For each example is reported the LLD query used to generate it. Visualizations include top 5 boxes above 0.3 confidence. For images where both QMD and the object detector provide a detection we report the detection score for both models.}
\label{fig:viz_lld}
\end{figure*}
In Fig.~\ref{fig:viz_lld} we provide visualizations for the COCO models reported in Tab.~4 of the paper. By modulating the detection using both the label and the coarse location, QMD can better localize difficult objects with respect to the pipeline with a standard detector followed by post-processing. This results in an improvement of +12.7\% LLD AP@0.5 for OID, and +4.8\% LLD AP@[.5:.95] for COCO (Sec.~4.8 of the paper).

One limitation of QMD is that it can also seldom output boxes outside the region of interest.  I.e. different from post-processing, QMD does not always 100\% suppress the confidence scores for boxes outside the region of interest. The confidence for such boxes is normally low. This can be seen for the ``Right Cow'' query in Fig.~\ref{fig:viz_lld}.

\section{Class-agnostic QMD}
In this Section we provide additional results for a class-agnostic version of multi-task QMD. When using the $[1, 1, 1, ..., 1]$ query, class-agnostic QMD behaves as a standard class-agnostic detector, and is thus unable to tell apart objects from different classes. On the other hand, by requesting a given label, it is possible to condition the model to produce ``class-agnostic'' boxes only for the desired label.

\begin{table}[h]
\footnotesize
    \centering
    \begin{tabularx}{\linewidth}{|p{0.5cm}p{1.2cm}p{0.8cm}|p{1.0cm}|p{1.0cm}|X|}
        \hline
        Dataset & Backbone & Model & SLD AP & KLD AP & DET AP\\
        \hline
        OID & MobileNetV2 & Detector & 51.3 & 45.4 & 41.7 \\
        OID & MobileNetV2 & QMD & \textbf{67.2} & \textbf{60.1} & \textbf{46.4} \\
        \hline
        COCO & MobileNetV2 & Detector & 28.9 & 25.6 & 25.7  \\
        COCO & MobileNetV2 & QMD & \textbf{33.3} & \textbf{28.7} & \textbf{27.3} \\
        \hline
        COCO & ResNet101 & Detector &  47.7 & 44.0 & 43.4 \\
        COCO & ResNet & QMD & \textbf{50.5} & \textbf{46.7} & \textbf{44.5} \\
        \hline
    \end{tabularx}
    \caption{\textbf{Class-agnostic QMD on OID and COCO}.  For SLD / KLD the model is asked to put boxes on objects belonging to specific label(s) in each image. DET AP refers to the class-agnostic detection AP.}
    \label{tab:class_agnostic_results}
\end{table}
In Tab.~\ref{tab:class_agnostic_results} we report the results for class-agnostic QMD on COCO and OID v4. By adopting a simple multi-task training approach, the class-agnostic detection AP achieved by QMD is always the highest in the benchmark.
Similarly to the multi-class results (Sec. 4.6 in the submitted manuscript), we observe much larger gains on OID v4 (Tab.~\ref{tab:class_agnostic_results} above) WRT COCO (Tab. 2 on the submitted manuscript). We believe this is due to the larger number of classes in OID.

To summarize, the experiment in Tab.~\ref{tab:class_agnostic_results}  shows that query-modulation could also be used to strongly condition the class-agnostic box-proposal stage for two-stage object detectors.

\subsection{Visualizations on COCO}\label{sec:experiments_query_modulated_detector}
\begin{figure*}[ht]
\centering
\includegraphics[width=\textwidth]{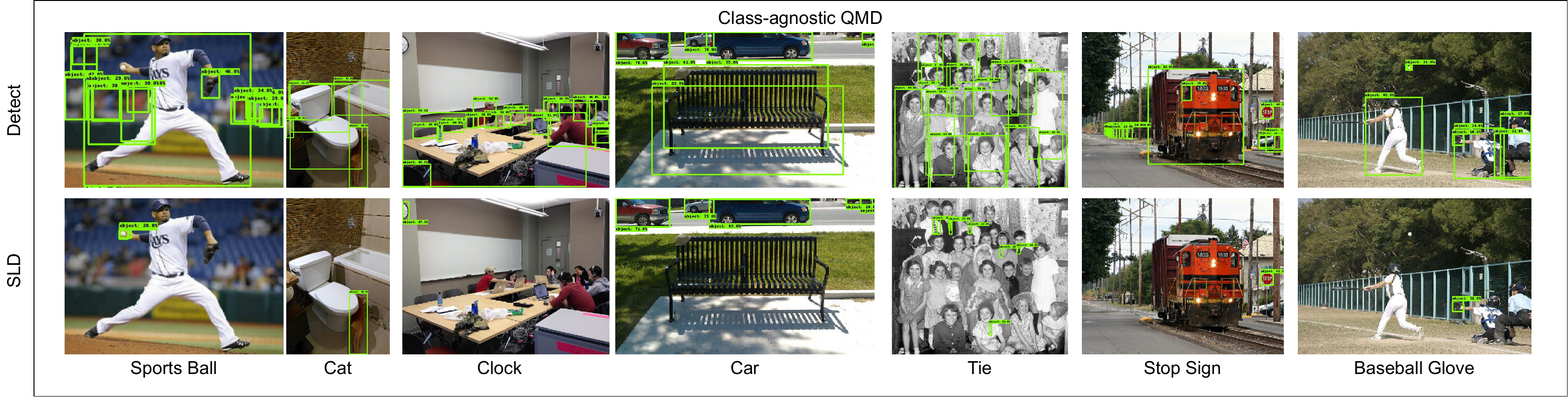}
\caption{Visualization of results for the MobileNetV2 class-agnostic QMD on COCO. First row: class-agnostic detection results. Second row: SLD results, with the corresponding query. Both rows are obtained using class-agnostic QMD.}
\label{fig:viz_class_agnostic_qmd}
\end{figure*}

In Fig.~\ref{fig:viz_class_agnostic_qmd} we visualize outputs of class-agnostic QMD, on COCO-valid. Standard class-agnostic detection results are visible in the first row. The second row shows how, by requesting a given label, it is possible to condition class-agnostic QMD to produce ``class-agnostic'' boxes only for the desired label.

In Fig.~\ref{fig:viz_top_classes} we provide some visualizations for class-agnostic QMD, compared to multiclass detection with post processing. In many cases (Refrigerator, Teddy Bear, Horse, Bed) the object cannot be obtained by post-processing the multi-class detector, as the latter does not provide a detection for the desired class. On the other hand, in cases where objects similar to the queried label are present in the image, QMD can provide false-positive detections on the distractor objects.


\subsection{Per-class results on COCO}
In Fig.~\ref{fig:viz_per_class_ap} and Tab.~\ref{tab:qmd_per_class_ap} we report the COCO-valid class-by-class AP achieved by:
\begin{itemize}
    \item Class-agnostic QMD, with each class queried only on images actually containing that class.
    \item Multi-class detector post-processed by pruning out detections for classes not appearing in a given image.
    \item Multi-class detector as is, reported for reference.
\end{itemize}

Overall, QMD improves over the post-processing by 4.1\%. QMD also provides the highest AP for any given class, except Snowboard. QMD provides the highest average improvement WRT the multi-class detector (6.8\% = 4.1\% + 2.7\%).

\begin{figure*}[ht]
\centering
\includegraphics[width=\textwidth]{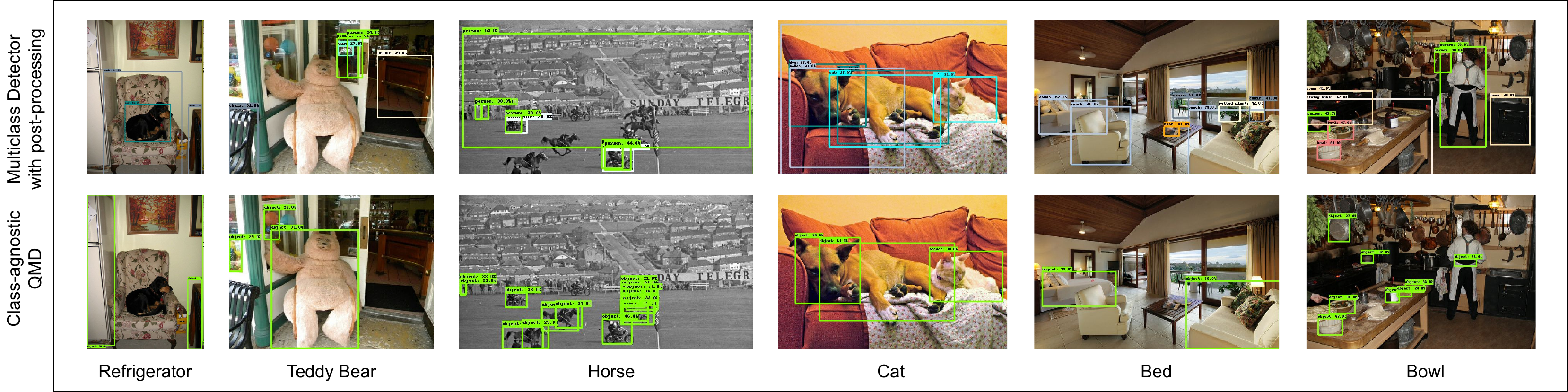}
\caption{Eight top-scoring detections above 0.2 confidence, on COCO-valid. First row: multiclass detector with post-proc, pruning out all detections for classes not in the image. Second row class-agnostic QMD with the associated query. All models use a MobileNetV2 feature extractor.}
\label{fig:viz_top_classes}
\end{figure*}
\begin{figure*}[h]
\centering
\includegraphics[width=\textwidth]{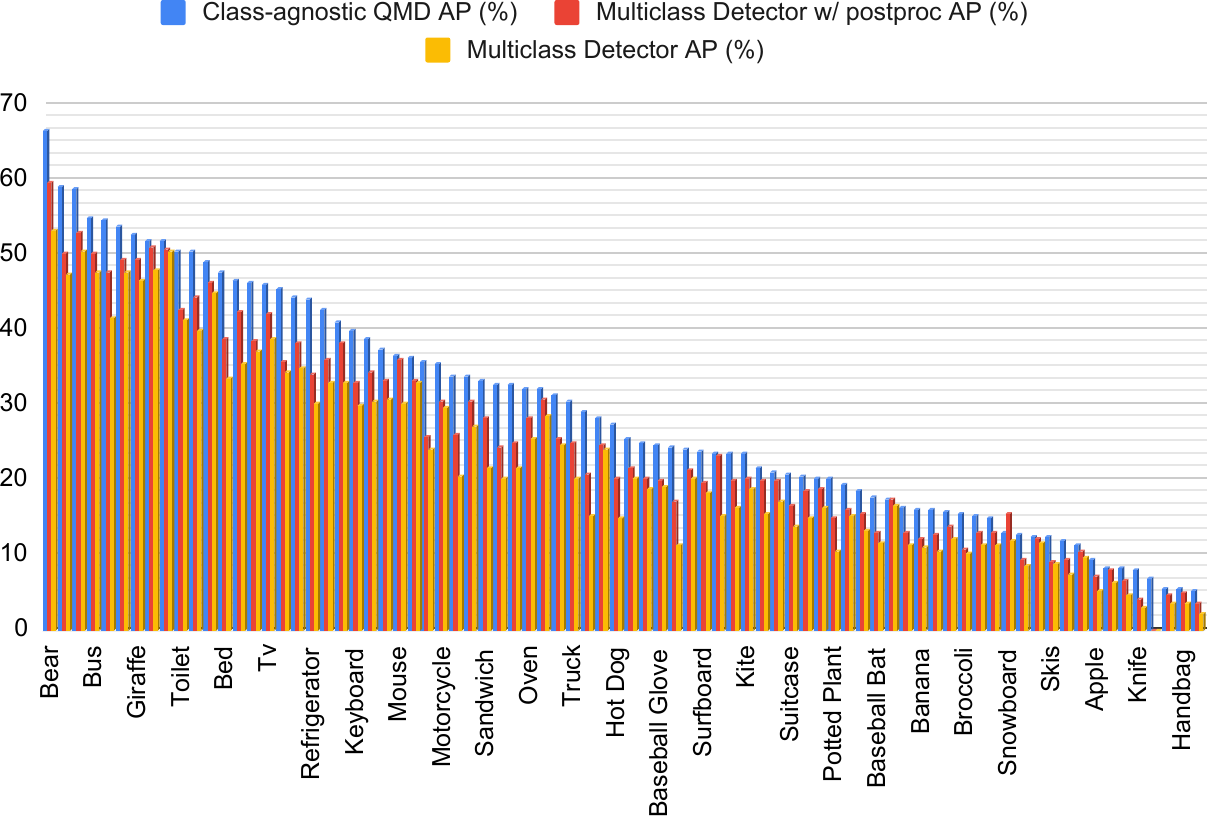}
\caption{Visualizations of per-class AP obtained by, respectively: class-agnostic QMD; the-multi-class detector post-processed to retain only detections for classes that do appear in the image; the multi-class detector as is. Classes are sorted by the AP obtained by QMD.
On average QMD improves multi-class detection results by 6.8\%, while post-processing only improves AP by 2.7\%. All models use a MobileNetV2 feature extractor.}
\label{fig:viz_per_class_ap}
\end{figure*}

\begin{table*}[ht]
\tiny
\begin{tabular}{lccc|rr}
Class          & Class-agnostic QMD AP (\%) & Multiclass Detector w/ postproc AP (\%)&  Multiclass Detector AP (\%) & (QMD $-$ Multiclass w/ postproc) AP Delta (\%) &(Multiclass w/ postproc $-$ Multiclass) AP Delta (\%) \\
\hline
Refrigerator   & 44                       & 34              & 30.1            & 10                            & 3.9                                   \\
Teddy Bear     & 35.7                     & 25.8            & 24              & 9.9                           & 1.8                                   \\
Horse          & 45.6                     & 35.8            & 34.4            & 9.8                           & 1.4                                   \\
Cat            & 59.2                     & 50.1            & 47.3            & 9.1                           & 2.8                                   \\
Bed            & 47.7                     & 38.9            & 33.5            & 8.8                           & 5.4                                   \\
Bowl           & 32.8                     & 24.4            & 20.2            & 8.4                           & 4.2                                   \\
Cake           & 29.1                     & 20.9            & 15.1            & 8.2                           & 5.8                                   \\
Toilet         & 50.6                     & 42.8            & 41.4            & 7.8                           & 1.4                                   \\
Laptop         & 46.4                     & 38.6            & 37.2            & 7.8                           & 1.4                                   \\
Donut          & 32.6                     & 24.9            & 21.6            & 7.7                           & 3.3                                   \\
Dining Table   & 33.8                     & 26.1            & 20.5            & 7.7                           & 5.6                                   \\
Hot Dog        & 27.5                     & 20.2            & 14.8            & 7.3                           & 5.4                                   \\
Scissors       & 24.5                     & 17.2            & 11.3            & 7.3                           & 5.9                                   \\
Dog            & 54.6                     & 47.6            & 41.7            & 7                             & 5.9                                   \\
Hair Drier     & 7                        & 0               & 0               & 7                             & 0                                     \\
Keyboard       & 39.9                     & 32.9            & 29.9            & 7                             & 3                                     \\
Bear           & 66.6                     & 59.7            & 53.2            & 6.9                           & 6.5                                   \\
Couch          & 42.7                     & 36.1            & 32.9            & 6.6                           & 3.2                                   \\
Fire Hydrant   & 50.4                     & 44.3            & 39.8            & 6.1                           & 4.5                                   \\
Pizza          & 44.4                     & 38.4            & 34.9            & 6                             & 3.5                                   \\
Train          & 58.9                     & 52.9            & 50.6            & 6                             & 2.3                                   \\
Skateboard     & 31.4                     & 25.6            & 24.7            & 5.8                           & 0.9                                   \\
Truck          & 30.4                     & 25              & 20.3            & 5.4                           & 4.7                                   \\
Potted Plant   & 20.1                     & 14.8            & 10.5            & 5.3                           & 4.3                                   \\
Sandwich       & 33.3                     & 28.2            & 21.6            & 5.1                           & 6.6                                   \\
Motorcycle     & 35.4                     & 30.4            & 29.7            & 5                             & 0.7                                   \\
Baseball Glove & 24.7                     & 19.8            & 19              & 4.9                           & 0.8                                   \\
Baseball Bat   & 17.8                     & 13              & 11.5            & 4.8                           & 1.5                                   \\
Broccoli       & 15.4                     & 10.7            & 10.1            & 4.7                           & 0.6                                   \\
Umbrella       & 25                       & 20.3            & 18.9            & 4.7                           & 1.4                                   \\
Bus            & 54.9                     & 50.2            & 47.7            & 4.7                           & 2.5                                   \\
Cow            & 38.9                     & 34.3            & 30.6            & 4.6                           & 3.7                                   \\
Airplane       & 53.7                     & 49.4            & 47.8            & 4.3                           & 1.6                                   \\
Suitcase       & 20.9                     & 16.6            & 13.8            & 4.3                           & 2.8                                   \\
Frisbie        & 46.5                     & 42.3            & 35.5            & 4.2                           & 6.8                                   \\
Clock          & 37.3                     & 33.2            & 30.9            & 4.1                           & 2.3                                   \\
Surfboard      & 23.7                     & 19.7            & 18.2            & 4                             & 1.5                                   \\
Oven           & 32.2                     & 28.3            & 25.5            & 3.9                           & 2.8                                   \\
Tv             & 46                       & 42.1            & 38.8            & 3.9                           & 3.3                                   \\
Banana         & 16.1                     & 12.3            & 11.1            & 3.8                           & 1.2                                   \\
Knife          & 8                        & 4.2             & 3               & 3.8                           & 1.2                                   \\
Sink           & 25.4                     & 21.6            & 20.3            & 3.8                           & 1.3                                   \\
Cup            & 23.5                     & 19.8            & 16.2            & 3.7                           & 3.6                                   \\
Tennis Racket  & 28.4                     & 24.7            & 24              & 3.7                           & 0.7                                   \\
Bottle         & 16.1                     & 12.6            & 10.4            & 3.5                           & 2.2                                   \\
Sheep          & 33.8                     & 30.4            & 27.1            & 3.4                           & 3.3                                   \\
Giraffe        & 52.6                     & 49.3            & 46.7            & 3.3                           & 2.6                                   \\
Chair          & 16.2                     & 12.9            & 11.3            & 3.3                           & 1.6                                   \\
Bicycle        & 19.4                     & 16.1            & 15.2            & 3.3                           & 0.9                                   \\
Bench          & 18.6                     & 15.4            & 13.3            & 3.2                           & 2.1                                   \\
Skis           & 12.4                     & 9.2             & 8.8             & 3.2                           & 0.4                                   \\
Kite           & 23.5                     & 20.3            & 18.9            & 3.2                           & 1.4                                   \\
Person         & 36.4                     & 33.3            & 33              & 3.1                           & 0.3                                   \\
Fork           & 12.6                     & 9.5             & 8.5             & 3.1                           & 1                                     \\
Microwave      & 41.1                     & 38.3            & 33              & 2.8                           & 5.3                                   \\
Elephant       & 49                       & 46.3            & 44.9            & 2.7                           & 1.4                                   \\
Car            & 24                       & 21.3            & 20.3            & 2.7                           & 1                                     \\
Remote         & 11.8                     & 9.3             & 7.5             & 2.5                           & 1.8                                   \\
Apple          & 9.3                      & 7.2             & 5.2             & 2.1                           & 2                                     \\
Wine Glass     & 15.1                     & 13              & 11.3            & 2.1                           & 1.7                                   \\
Tie            & 15.7                     & 13.7            & 12.2            & 2                             & 1.5                                   \\
Sports Ball    & 20.5                     & 18.6            & 14.9            & 1.9                           & 3.7                                   \\
Carrot         & 14.9                     & 13.1            & 11.3            & 1.8                           & 1.8                                   \\
Vase           & 21.6                     & 19.9            & 15.6            & 1.7                           & 4.3                                   \\
Spoon          & 5.2                      & 3.6             & 2.1             & 1.6                           & 1.5                                   \\
Backpack       & 8.2                      & 6.7             & 4.7             & 1.5                           & 2                                     \\
Bird           & 20.1                     & 18.7            & 16.4            & 1.4                           & 2.3                                   \\
Parking Meter  & 32.1                     & 30.9            & 28.6            & 1.2                           & 2.3                                   \\
Cellphone      & 21                       & 19.9            & 17.2            & 1.1                           & 2.7                                   \\
Zebra          & 51.9                     & 50.8            & 50.4            & 1.1                           & 0.4                                   \\
Book           & 5.6                      & 4.6             & 3.6             & 1                             & 1                                     \\
Stop Sign      & 51.9                     & 50.9            & 48              & 1                             & 2.9                                   \\
Traffic Light  & 11.2                     & 10.5            & 9.7             & 0.7                           & 0.8                                   \\
Mouse          & 36.6                     & 36              & 30.1            & 0.6                           & 5.9                                   \\
Toaster        & 23.6                     & 23.2            & 15.1            & 0.4                           & 8.1                                   \\
Handbag        & 5.4                      & 5               & 3.5             & 0.4                           & 1.5                                   \\
Toothbrush     & 8.4                      & 8.1             & 6.2             & 0.3                           & 1.9                                   \\
Boat           & 12.5                     & 12.3            & 11.5            & 0.2                           & 0.8                                   \\
Orange         & 17.5                     & 17.4            & 16.6            & 0.1                           & 0.8                                   \\
Snowboard      & 13                       & 15.5            & 11.9            & -2.5                          & 3.6                                   \\
\hline
Mean           & 29.4                     & 25.3            & 22.6            & 4.1                           & 2.7                                  
\end{tabular}
\caption{\textbf{Per-class localization AP on COCO-valid.} For QMD, each class is queried only on images actually containing it. For the multiclass detector with post-processing, predictions for each class are retained only on images actually containing it. All models use a MobileNetV2 feature extractor.}
\label{tab:qmd_per_class_ap}
\end{table*}

\end{document}